# Cascades of Regression Tree Fields for Image Restoration

Uwe Schmidt, *Student Member, IEEE,* Jeremy Jancsary, *Member, IEEE,* Sebastian Nowozin, Stefan Roth, *Member, IEEE,* and Carsten Rother, *Member, IEEE*

**Abstract**—Conditional random fields (CRFs) are popular discriminative models for computer vision and have been successfully applied in the domain of image restoration, especially to image denoising. For image deblurring, however, discriminative approaches have been mostly lacking. We posit two reasons for this: First, the blur kernel is often only known at test time, requiring any discriminative approach to cope with considerable variability. Second, given this variability it is quite difficult to construct suitable features for discriminative prediction. To address these challenges we first show a connection between common half-quadratic inference for generative image priors and Gaussian CRFs. Based on this analysis, we then propose a cascade model for image restoration that consists of a Gaussian CRF at each stage. Each stage of our cascade is semi-parametric, *i.e.* it depends on the instance-specific parameters of the restoration problem, such as the blur kernel. We train our model by loss minimization with synthetically generated training data. Our experiments show that when applied to non-blind image deblurring, the proposed approach is efficient and yields state-of-the-art restoration quality on images corrupted with synthetic and real blur. Moreover, we demonstrate its suitability for image denoising, where we achieve competitive results for grayscale and color images.

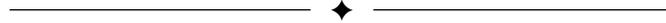

## 1 INTRODUCTION

IMAGE restoration is an important and long-studied field, manifesting itself in numerous applications, such as image denoising, deblurring, or super-resolution. Image restoration can be seen as an *inverse problem*, where an image corruption process – modeled by a data (or likelihood) term – is to be inverted. Such an inversion is typically mathematically ill-posed, which necessitates the use of regularization (or prior knowledge).

Prior knowledge can be imposed in a variety of ways. Discriminative approaches have received increasing attention in recent years, particularly for image denoising [1, 2, 3, 4], where they often yield state-of-the-art restoration performance combined with low computational effort. In other image restoration applications, such as non-blind image deblurring [5, 6, 7] on the other hand, generative approaches are standard. We argue that the lack of discriminative methods for these applications stems from their more challenging data term with additional instance-specific parameters, which are necessary to capture the image corruption process properly. In non-blind deblurring[1], for example, not only the noise level, but also the blur kernel has to be given to the algorithm,
which is often only known at test time and may vary from (image) instance to instance. For deblurring the instance-specific parameters thus correspond to the blur kernel. In a discriminative approach it is, however, quite difficult to cope with such instance-specific parameters.

In this paper we introduce a *discriminative* image restoration approach for applications that can be expressed via arbitrary quadratic data terms (Gaussian likelihoods). The first major challenge we address is that the number of possible inputs to such a model increases exponentially with the number of (input) parameters. Because of that, training a conditional model for every possible instance-specific parameter, *e.g.* for every possible blur kernel [8], is very costly or even infeasible. In this paper, we thus argue that it is important to be able to train a *single* model that outputs a restored image given an arbitrary input image *and* instance parameter, such as the blur kernel. We address the challenge of capturing the input distribution variability by using a semi-parametric approach: We specify part of the model explicitly by means of instance-specific parameters and capture the remaining variability using non-parametric regression trees. As a consequence, we assume access to these instance-specific parameters during training and testing, for example by means of an estimate. More specifically, our approach is based on regression tree fields (RTFs) [9], a Gaussian conditional random field (CRF) in which the parameters of the Gaussian field are determined through regression trees.

When considering image deblurring in contrast to image denoising, a second major challenge arises: The great variability of the image corruption due to blur that is only known at test time makes it also rather

- U. Schmidt and S. Roth are with the Department of Computer Science, TU Darmstadt, Darmstadt, Germany.
- C. Rother is with the Department of Computer Science, TU Dresden, Dresden, Germany.
- S. Nowozin is with Microsoft Research Ltd, Cambridge, UK.
- J. Jancsary is with Nuance Communications, Vienna, Austria. This work was done while at Microsoft Research Ltd, Cambridge, UK.

1. A more precise term would be *deconvolution* instead of deblurring when a stationary blur assumption is made. We use the more general terminology as our discussion is not limited to deconvolution.

difficult to derive suitable features from the input image, which are then used as inputs for the regression trees. To address this we take inspiration from common half-quadratic approaches to image restoration [10, 11, 12]. In particular, we observe that while half-quadratic MAP estimation makes its final prediction also based on a Gaussian random field, the parameters of this random field are iteratively refined during the inference procedure. This is in contrast to typical Gaussian CRF approaches, where the parameters are estimated in a one-shot fashion. Motivated by that, we introduce a model cascade based on regression tree fields. The first stage predicts a relatively crude estimate that removes dominant image blur, which is however very useful to define better input features for later stages. In this way the deblurred image is incrementally refined in each stage. We apply our discriminative prediction cascade also to the problem of image denoising, where we find that the cascade architecture benefits image quality as well, albeit somewhat less than for deblurring.

Our model cascade is trained discriminatively by minimizing an application-specific loss function (here, PSNR) on a training set. To make this feasible, we synthesize training data according to the given application-specific data term. One challenge in case of deblurring is that sufficient training data must be available for discriminative training, but realistic image blurs are quite scarce [13, 14]. To overcome this limitation, we use synthetically generated blur kernels based on a simple motion model, which we show to generalize well to kernels encountered in practice.

**Contributions.** This paper makes the following contributions: *(1)* We analyze commonly used half-quadratic regularization [10, 11] with sparse image priors, and draw connections to discriminative prediction with a CRF; *(2)* we introduce a discriminative prediction cascade for image restoration based on regression tree fields, which naturally arises as a generalization of half-quadratic inference; *(3)* we employ a semi-parametric approach at each prediction stage, which allows a single trained model to cope with parameters that vary from instance to instance, such as the blur kernel in image deblurring; *(4)* we train our model with data that was obtained by using realistic, but synthetically generated blur kernels and experimentally demonstrate that the trained model generalizes to unseen real blur kernels; *(5)* we demonstrate state-of-the-art performance on a synthetically blurred test set [7] and on two realistic data sets for camera shake [14, 15]. While previous non-blind deblurring approaches have for the most part either been very fast but with inferior performance, or slow but with high-quality results [*e.g.* 7], our approach delivers state-of-the-art deblurring performance with an efficient inference method that allows deblurring even higher-resolution images; *(6)* we demonstrate state-of-the-art performance for a (grayscale) image denoising benchmark. We also train our model for color denoising and show its superiority over applying our grayscale denoising model independently for each color channel.

This paper is an extended version of [16].

**Background in image deblurring.**[2] Image blur (*e.g.*, camera shake) is one of the main sources of image corruption in digital photography and hard to undo. Image deblurring has thus been an active area of research, starting with the pioneering work of Lucy [17] and Richardson [18]. Recent work has predominantly focused on *blind deblurring* [*e.g.* 15, 19, 20, 21], particularly on estimating the blur from images (stationary and non-stationary [22]). However, relatively little attention has been paid to *non-blind deblurring*, that is, restoring the image given known or estimated image blur. Yet, this is an important problem since most blind deblurring approaches separate the problem into blur estimation and non-blind deblurring (theoretically justified by Levin *et al*. [13]). Furthermore, the task of non-blind deconvolution is prevalent for microscopic images, since the point spread function of the microscope can typically by measured accurately. To this date, most approaches rely on the classical Lucy-Richardson algorithm as non-blind deblurring component [*e.g.*, 19], or use manually-defined image priors formulated as Markov random fields (MRFs) with sparse, *i.e.* non-Gaussian, potential functions [5, 6, 21]. Learning-based approaches have been restricted to generatively trained models [7], but have found limited adoption due to computational challenges from inference. In this paper we assume stationary image blur, *i.e.* the observed image is the result of convolving the unknown original image with a blur kernel (+ noise), but our approach is not limited to this setup and can be extended to non-uniform image blurs.

## 2 Generalizing Half-quadratic Regularization

To motivate our discriminative approach and understand its connections to the existing literature, it is beneficial to recall half-quadratic regularizers [10, 11, 12] and their relation to recent image restoration approaches. In image deblurring, denoising and other restoration applications, sparse image priors are frequently used for regularization [*e.g.* 5, 6, 23]. Typically, they model an image $\mathbf{x}$ through the response of linear filters $\mathbf{f}_j$ (*e.g.*, horizontal and vertical image derivatives), which induce overlapping cliques $c \in \mathcal{C}_j$ in the corresponding Markov random field (MRF) prior:

$$p(\mathbf{x}) \propto \prod_j \prod_{c \in \mathcal{C}_j} \rho_j(\mathbf{f}_j^\mathrm{T} \mathbf{x}_{(c)}). \qquad (1)$$

A sparse (non-Gaussian) potential function $\rho_j$ models the filter response of $\mathbf{f}_j$ to the clique pixels $\mathbf{x}_{(c)}$.

The image corruption process is often modeled by specifying a Gaussian likelihood $p(\mathbf{y}|\mathbf{x}) = \mathcal{N}(\mathbf{y}; \mathbf{K}\mathbf{x}, \sigma^2 \mathbf{I})$

---

2. To improve readability, related work will be reviewed in-line.

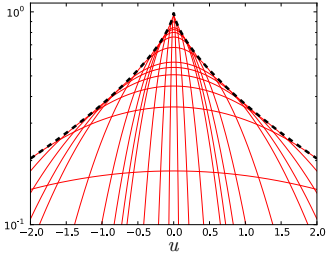

Fig. 1. **Half-quadratic representation of a sparse potential.** Hyper-Laplacian $\rho(u) = \exp(-|u|^\gamma)$ (dashed, black), where $\rho(u) = \sup_z \phi(u,z)$, and $\phi(u,z) = \exp(-\frac{1}{2}u^2 z - \psi_m(z))$ (solid, red) with $\psi_m(z) = (1-\gamma/2) \cdot (z/\gamma)^{\frac{\gamma}{\gamma-2}}$ and $\gamma = 2/3$.

for the observed, corrupted image $\mathbf{y}$. In the case of non-blind deconvolution, we have $\mathbf{Kx} \equiv \mathbf{k} \otimes \mathbf{x}$, where $\mathbf{K}$ is the blur matrix that corresponds to convolving the image with a blur kernel $\mathbf{k}$. The image noise is assumed to be additive white Gaussian noise with variance $\sigma^2$. The problem of image denoising arises as a special case with $\mathbf{K} = \mathbf{I}$ being the identity matrix. If multiplication with $\mathbf{K}$ reduces the spatial resolution of the image, the likelihood models the problem of super-resolution. Using Bayes' theorem, one obtains the posterior distribution over the restored image as $p(\mathbf{x}|\mathbf{y}) \propto p(\mathbf{y}|\mathbf{x}) \cdot p(\mathbf{x})$.

The principle of half-quadratic regularization [10, 11, 12] is to ease inference (*e.g.*, MAP estimation) by introducing (independent) auxiliary/latent variables $z_{jc}$ for each filter and image clique, such that the prior is retained by performing an operation $\bigoplus \in \{\max, \sup, \sum, \int, \ldots\}$ that eliminates the auxiliary variables:

$$p(\mathbf{x}) \propto \prod_j \prod_{c \in \mathcal{C}_j} \bigoplus_{z_{jc}} \phi_j(\mathbf{f}_j^T \mathbf{x}_{(c)}, z_{jc}). \tag{2}$$

Assuming that $\bigoplus$ commutes with the product operation, which is the case by choice, we can define an augmented prior as

$$p(\mathbf{x}, \mathbf{z}) \propto \prod_j \prod_{c \in \mathcal{C}_j} \phi_j(\mathbf{f}_j^T \mathbf{x}_{(c)}, z_{jc}) \tag{3}$$

with

$$p(\mathbf{x}) \propto \bigoplus_{\mathbf{z}} p(\mathbf{x}, \mathbf{z}). \tag{4}$$

Two choices for the form of $\phi_j$ are prevalent: the multiplicative form $\phi_j(u,z) = \exp\left(-\frac{1}{2}u^2 z - \psi_m(z)\right)$ [11] and the additive form $\phi_j(u,z) = \exp\left(-b(u-z)^2 - \psi_a(z)\right)$ [10]. In either case, $\psi_m(z)$ respectively $\psi_a(z)$ (and $b$) are chosen such that $\rho_j(u) = \bigoplus_z \phi_j(u,z)$ (see example in Fig. 1). The name stems from the fact that the energy of $\phi_j(u,z)$ is quadratic in $u$ for a given value of $z$. This further implies that for a fixed setting of $\mathbf{z}$ the distribution $p(\mathbf{x}|\mathbf{z}) = \mathcal{N}(\mathbf{x}; \boldsymbol{\mu}_{\mathbf{x}|\mathbf{z}}, \boldsymbol{\Sigma}_{\mathbf{x}|\mathbf{z}})$ is jointly Gaussian. When combined with a Gaussian likelihood, we obtain a Gaussian posterior for a fixed setting of $\mathbf{z}$:

$$p(\mathbf{x}|\mathbf{y},\mathbf{z}) \propto \mathcal{N}(\mathbf{y}; \mathbf{Kx}, \sigma^2 \mathbf{I}) \cdot \mathcal{N}(\mathbf{x}; \boldsymbol{\mu}_{\mathbf{x}|\mathbf{z}}, \boldsymbol{\Sigma}_{\mathbf{x}|\mathbf{z}})$$
$$\propto \mathcal{N}(\mathbf{x}; \boldsymbol{\mu}_{\mathbf{x}|\mathbf{y},\mathbf{z}}, \boldsymbol{\Sigma}_{\mathbf{x}|\mathbf{y},\mathbf{z}}). \tag{5}$$

The benefit of this is that MAP estimation can now be carried out on the augmented posterior $p(\mathbf{x}, \mathbf{z}|\mathbf{y})$ with a variational EM algorithm [*cf*. 24, 25] that alternates between maximizing $p(\mathbf{x}|\mathbf{y},\mathbf{z})$ and using $p(\mathbf{z}|\mathbf{x},\mathbf{y})$ to update the auxiliary variables; the type of update depends on the choice of the operation $\bigoplus$. Maximizing $p(\mathbf{x}|\mathbf{y},\mathbf{z})$ w.r.t. $\mathbf{x}$ amounts to computing $\boldsymbol{\mu}_{\mathbf{x}|\mathbf{y},\mathbf{z}}$, which requires solving a sparse linear equation system based on the easily accessible precision matrix $\boldsymbol{\Sigma}_{\mathbf{x}|\mathbf{y},\mathbf{z}}^{-1}$; note that the covariance $\boldsymbol{\Sigma}_{\mathbf{x}|\mathbf{y},\mathbf{z}}$ cannot be used as it is not sparse. Updating $\mathbf{z}$ based on $p(\mathbf{z}|\mathbf{y},\mathbf{x})$ is easy, because it can be done for each scalar variable $z_{jc}$ individually (*e.g.*, with a table lookup), since all $z_{jc}$ are independent:

$$p(\mathbf{z}|\mathbf{x},\mathbf{y}) \propto \prod_j \prod_{c \in \mathcal{C}_j} p(z_{jc}|\mathbf{x},\mathbf{y}) \tag{6}$$

$$p(z_{jc}|\mathbf{x},\mathbf{y}) \propto \phi_j(\mathbf{f}_j^T \mathbf{x}_{(c)}, z_{jc}). \tag{7}$$

By using the fact that a wide variety of robust (sparse) potentials $\rho_j$ can be expressed (or approximated) by taking the supremum over the auxiliary variables $\mathbf{z}$ [26], one can re-formulate the majority of sparse image priors in this way. Levin *et al.* [5] and Krishnan and Fergus [6] have employed this principle for efficient image deblurring. Note that MRF image priors based on Gaussian scale mixtures (GSMs) [27] can also be interpreted as an instance of half-quadratic regularization in which $\bigoplus = \sum$ (or $\bigoplus = \int$ for infinite GSMs). This has been used by Schmidt *et al.* for image denoising [28] and deblurring [7] with sampling-based inference, which alternates between sampling from $p(\mathbf{x}|\mathbf{y},\mathbf{z})$ and $p(\mathbf{z}|\mathbf{x},\mathbf{y})$. Babacan *et al.* [29] have exploited half-quadratic representations in the context of blind deconvolution.

### 2.1 Discriminative alternative

To see how classical half-quadratic regularization can be connected to a discriminative approach, it is instructive to consider what happens during the last inference iteration. Once the final set of latent variables $\mathbf{z}^*$ has been determined from Eq. (7), the output image $\mathbf{x}^*$ is inferred by maximizing $p(\mathbf{x}|\mathbf{y},\mathbf{z}^*)$ from Eq. (5). This distribution is nothing but an anisotropic (or inhomogeneous) Gaussian random field, whose mean and covariance depend on $\mathbf{y}$ and $\mathbf{z}^*$ (and also $\mathbf{K}$ and $\sigma$).

Therefore $\boldsymbol{\mu}_{\mathbf{x}|\mathbf{y},\mathbf{z}^*}$ and $\boldsymbol{\Sigma}_{\mathbf{x}|\mathbf{y},\mathbf{z}^*}$ are the mean and covariance parameters of a multivariate normal distribution defined on the whole image, chosen through $\mathbf{z}^*$ so as to hopefully lead to good restoration results. The value of $\mathbf{z}^*$ depends on the specific choice of potential functions $\rho_j$ and their half-quadratic representations $\phi_j$ (Eq. 7).

It is now natural to ask whether we can instead directly regress the Gaussian random field parameters from the input image. More specifically, we can regress a precision matrix $\boldsymbol{\Theta}(\mathbf{y})$ and a vector $\boldsymbol{\theta}(\mathbf{y})$, leading to $\boldsymbol{\mu} := [\boldsymbol{\Theta}(\mathbf{y})]^{-1}\boldsymbol{\theta}$ and $\boldsymbol{\Sigma} := [\boldsymbol{\Theta}(\mathbf{y})]^{-1}$. Then the mean $\boldsymbol{\mu}$ and the covariance $\boldsymbol{\Sigma}$ are learned functions of the observed image $\mathbf{y}$. There are three potential advantages: *First*, we avoid the expensive iterative computation of the half-quadratic optimization. *Second*, we can regress the parameters discriminatively in order to optimize a given performance measure, such as PSNR. *Third*, we are no longer constrained to the form of Eq. (7) so that we can now use an expressive regression model on the input image. That is, we are not restricting[3] the resulting model compared to Eq. (5); in fact, we can potentially learn a more expressive model.

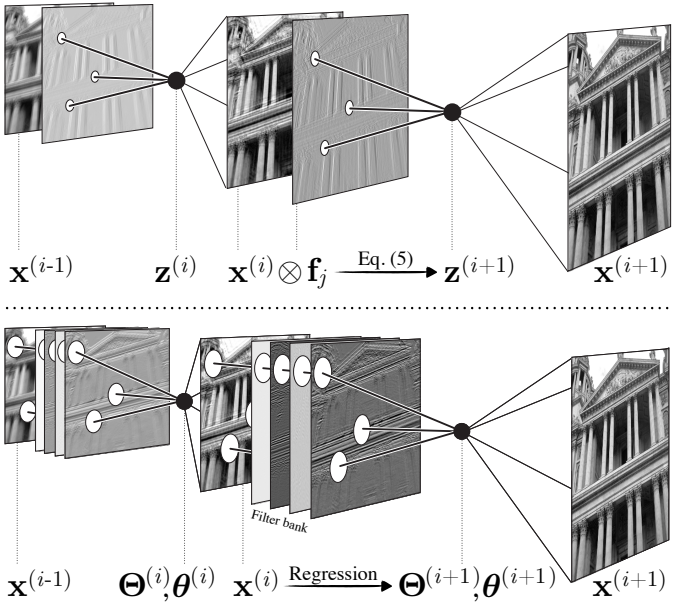

Fig. 2. **Standard half-quadratic vs. discriminative cascade.** In a standard half-quadratic approach *(top)*, each $z_{jc}$ can only be updated via Eq. (7) based on the filter response $\mathbf{f}_j^T \mathbf{x}_{(c)}$ of the pixels in the local clique (small white circles, only one filtered image $\mathbf{f}_j \otimes \mathbf{x}$ shown). In the proposed discriminative cascade *(bottom)*, one can use arbitrary features of the image over larger areas (large white circles) to find model parameters $\Theta^{(i)}$ and $\boldsymbol{\theta}^{(i)}$ via regression. At each stage, the functions $\Theta^{(i)}$ and $\boldsymbol{\theta}^{(i)}$ depend on $\mathbf{y}$ through features, such as filter bank responses, image intensities, as well as $\mathbf{x}^{(i)}$ from previous iterations (not shown).

We arrived at this model from a novel analysis of the half-quadratic approximation, but predicting the means and covariances of a Gaussian model has been done before: Gaussian conditional random fields, first proposed by Tappen *et al.* [1], have led to competitive results in image denoising. We build on the more recent regression tree fields (RTFs) by Jancsary *et al.* [3, 9].

**Going beyond denoising.** While such Gaussian CRFs have been successful for image denoising, we argue that applying them to other image restoration applications, such as non-blind image deblurring, is more challenging, since it is difficult to directly regress suitable model parameters. To illustrate this difficulty, let us assume that $\mathbf{f}_j$ are first-order derivative filters. Then, in the generative approach one can think of $z_{jc}$ as modulating pairwise potentials: reducing smoothness constraints in case of large image derivatives *of the output image* $\mathbf{x}$, and imposing smoothness otherwise. In other words, in the generative approach $\mathbf{z}$ determines the local model of the restored image $\mathbf{x}$. Both $\mathbf{x}$ and $\mathbf{z}$ are iteratively refined via half-quadratic inference. In a discriminative model we have access only to the corrupted image $\mathbf{y}$ in order to determine suitable local models.

---

3. Note that any multivariate Gaussian distribution can always be expressed as a product of unary and pairwise terms [*cf.* 30], because its exponent is the sum of a linear and quadratic form (*i.e.*, homogeneous polynomials of degrees 1 and 2, respectively). Hence, the final MAP estimate in half-quadratic regularization comes from a pairwise MRF *even if the corresponding sparse image prior models high-order interactions*. This does not mean, however, that high-order dependencies are ignored. They are only hidden in the estimate $\mathbf{z}^*$.

But in the case of deblurring, the image content in $\mathbf{y}$ is shifted and combined with other parts of the image, depending on an instance-specific blur kernel. This makes the choice of local models difficult. We believe this is one of the reasons why discriminative non-blind deblurring approaches had not been attempted before.

The situation is much easier for image denoising, since it is typically assumed that noise is additive and pixel-independent; hence, one can regress model parameters quite well by averaging values in a neighborhood around a pixel, or more generally by applying a set of filters whose responses provide discriminative features to regress model parameters [*cf.* 1, 3].

## 2.2 Discriminative model cascade

To build a discriminative model for deblurring, we draw inspiration from the iterative refinement of $\mathbf{z}$ in half-quadratic regularization. We start with an educated guess of the Gaussian model parameters, regressed from the input image, to obtain a restored image $\mathbf{x}^{(1)}$, which is less corrupted than the original input image. We can then use this as an intermediate result to help regress refined Gaussian model parameters, in order to obtain a better restored image $\mathbf{x}^{(2)}$, *etc.*, effectively obtaining a cascade of refined models. Note that this is a general approach that is not only applicable to image deblurring; other restoration tasks may also benefit from such a model cascade and repeated refinement of the auxiliary variables. As mentioned above, for the special case of image denoising, we can already obtain good parameters at the first model stage and thus obtain a high-quality initial result $\mathbf{x}^{(1)}$ whose restoration quality cannot be improved further as much as for more difficult problems, such as deblurring (*cf.* Sec. 4).

A key advantage of a discriminative approach for predicting model parameters $\Theta^{(i)}$, $\boldsymbol{\theta}^{(i)}$ at the $i^{\text{th}}$ stage is its flexibility. As discussed above, a standard generative half-quadratic approach updates each $z_{jc}$ only based on the local clique of the current estimate of the restored image (*cf.* Eq. 7). In a discriminative approach, we can regress the parameters based on arbitrary local and global properties of the input image as well as the current estimate of the restored image (see Fig. 2 for an illustration). Furthermore, we can use a specialized model (*i.e.*, regression function) for each stage, whereas an image prior in a generative approach does not change during inference. Consequently, we can expect to obtain better estimates in fewer iterations.

**Other related work.** This iterative procedure can also be linked with earlier ideas about iterative refinement. The idea of *auto-context* [31] is to use the same probabilistic model multiple times in sequence, where each model receives as input the observed image and the output of the previous model in the sequence. The proposed discriminative cascade is also related to the active random field of Barbu [2], which is a multi-stage approach for image denoising that is trained discriminatively. The key difference is that each stage in [2] corresponds to a gradient descent iteration of the model energy; moreover, the parameters are shared between all stages.

## 3 GAUSSIAN CRF FOR RESTORATION

As we have seen, a discriminative alternative to half-quadratic MAP estimation is conceptually attractive, but can be challenging due to the need of regressing local image models from the corrupted input image $\mathbf{y}$. To address this challenge we propose a novel Gaussian CRF $p(\mathbf{x}|\mathbf{y}; \mathbf{K})$ for image restoration with more challenging Gaussian image corruption models. Let us first consider non-blind image deblurring as a specific example. One challenge in devising such a model is that we cannot train a different model for every blur matrix $\mathbf{K}$; this difficulty may in fact be the reason why previous approaches require separate training for each specific blur kernel [8]. To see how this can be circumvented, we can take inspiration from generative approaches to deblurring and see how the Gaussian blur likelihood $p(\mathbf{y}|\mathbf{x}; \mathbf{K})$ contributes to the posterior distribution when assuming a Gaussian prior:

$$\begin{aligned}
p(\mathbf{x}|\mathbf{y}; \mathbf{K}) &\propto p(\mathbf{y}|\mathbf{x}; \mathbf{K}) \cdot p(\mathbf{x}) \\
&\propto \mathcal{N}(\mathbf{y}; \mathbf{Kx}, \mathbf{I}/\alpha) \cdot \mathcal{N}(\mathbf{x}; \boldsymbol{\Theta}^{-1}\boldsymbol{\theta}, \boldsymbol{\Theta}^{-1}) \\
&\propto \mathcal{N}\big(\mathbf{x}; (\alpha\mathbf{K}^\mathrm{T}\mathbf{K})^{-1}\alpha\mathbf{K}^\mathrm{T}\mathbf{y}, (\alpha\mathbf{K}^\mathrm{T}\mathbf{K})^{-1}\big) \\
&\quad \cdot \mathcal{N}(\mathbf{x}; \boldsymbol{\Theta}^{-1}\boldsymbol{\theta}, \boldsymbol{\Theta}^{-1}) \\
&\propto \mathcal{N}\big(\mathbf{x}; (\boldsymbol{\Theta}+\alpha\mathbf{K}^\mathrm{T}\mathbf{K})^{-1}(\boldsymbol{\theta}+\alpha\mathbf{K}^\mathrm{T}\mathbf{y}), \\
&\quad (\boldsymbol{\Theta}+\alpha\mathbf{K}^\mathrm{T}\mathbf{K})^{-1}\big),
\end{aligned} \quad (8)$$

where $\alpha = 1/\sigma^2$ relates to the noise level, $\boldsymbol{\Theta}$ is the precision of the Gaussian prior, and $\boldsymbol{\theta}$ relates to its mean. We can now define a Gaussian CRF in which the model parameters $\boldsymbol{\Theta}$ and $\boldsymbol{\theta}$ are not fixed, but regressed from the input image, *i.e.* $\boldsymbol{\Theta} \equiv \boldsymbol{\Theta}(\mathbf{y})$ and $\boldsymbol{\theta} \equiv \boldsymbol{\theta}(\mathbf{y})$ are functions of $\mathbf{y}$. Note that the CRF is parameterized by an instance-specific blur $\mathbf{K}$ as in Eq. (8); the blur is *not* used as an input feature to the regressor (although it could be).

Even though motivated through image deblurring, the proposed Gaussian CRF in Eq. (8) is not limited to this. Depending on the choice of the matrix $\mathbf{K}$, it can be used to model other applications, such as image super-resolution when $\mathbf{K}$ relates to a downsampling operation. A limitation of this construction is the assumption of Gaussian additive noise, which enables the combination of prior and likelihood terms in closed form.

For the problem of image denoising, *i.e.* $\mathbf{K} = \mathbf{I}$ is an identity matrix, it is worth noting that explicitly incorporating a component related to the likelihood as in Eq. (8) may not be necessary, since its contribution could be learned by the regression function. This approach has been pursued by previous work [1, 3, 9] and is also adopted here for the denoising experiments in Sec. 4. It also has the advantage of making no assumption about the type of noise corruption, which allows the removal of non-Gaussian noise, as shown by [3, 9]. In case of deblurring, however, our formulation in Eq. (8) does need to make a noise assumption, since a likelihood term is required to adapt the model to arbitrary blurs. But since the regression functions in our discriminative approach do not rely on a particular noise characteristic, our model can still cope with noise that violates the Gaussian assumption to some extent (see Sec. 4).

Once we have determined the parameters via regression, we can obtain a deblurred image as the MAP estimate, which can be derived in closed form as the mean of the Gaussian CRF,

$$\begin{aligned}
\arg\max_{\mathbf{x}} p(\mathbf{x}|\mathbf{y}; \mathbf{K}) = \\
(\boldsymbol{\Theta}(\mathbf{y}) + \alpha\mathbf{K}^\mathrm{T}\mathbf{K})^{-1}(\boldsymbol{\theta}(\mathbf{y}) + \alpha\mathbf{K}^\mathrm{T}\mathbf{y}),
\end{aligned} \quad (9)$$

and can be computed by solving a sparse linear system.

**Other related work.** In recent, independent work, Chen *et al.* [32] also combined a discriminatively-trained regularization term with an instance-specific data term for image deblurring and super-resolution. In contrast to our work, they do not provide a formal motivation and do not train the model specifically for these applications. Instead, they train their model for image denoising and then augment it with an instance-specific data term at test time. Furthermore, they cannot combine regularization and data terms in closed form, as they do not use Gaussian random fields. From a different point of view, Cho *et al.* [33] propose an adaptive prior for image restoration, which can be seen as a discriminative model whose parameters depend on the observed corrupted image. However, they do not attempt application-specific loss-based training, as we employ here.

### 3.1 Regression tree field

To make our approach concrete, we need to specify the regression functions $\boldsymbol{\Theta}(\mathbf{y})$ and $\boldsymbol{\theta}(\mathbf{y})$. To that end, we draw on the recently proposed *regression tree field* (RTF) model by Jancsary *et al.* [3, 9]. RTFs have shown state-of-the-art results in image restoration applications, such as image denoising, inpainting, and colorization.

In general, RTFs take the form of a Gaussian CRF in which a non-linear regressor is used to specify the local model parameters. Specifically, regression trees are used, where each leaf stores an individual linear regressor that determines a local potential. Since any Gaussian CRF can be decomposed into factors relating no more than two pixels, our posterior density attains the following form:

$$p(\mathbf{x}|\mathbf{y};\mathbf{K}) \propto \mathcal{N}(\mathbf{y};\mathbf{K}\mathbf{x},\mathbf{I}/\alpha) \cdot \prod_{j=1}^{J+1} \prod_{c \in \mathcal{C}_j} \phi_j(\mathbf{x}_{(c)},\mathbf{y}) \quad (10)$$

$$\phi_j(\mathbf{x}_{(c)},\mathbf{y}) \propto \exp\left(-\frac{1}{2}\mathbf{x}_{(c)}^\mathsf{T}\boldsymbol{\Theta}_c^j(\mathbf{y})\mathbf{x}_{(c)} + \mathbf{x}_{(c)}^\mathsf{T}\boldsymbol{\theta}_c^j(\mathbf{y})\right),$$

where $\mathcal{C}_j$ denotes all pairs of neighboring pixels in the $j^{\text{th}}$ of $J$ possible spatial configurations. Concretely, we use 8- and 24-neighborhoods depending on the application and stage in our prediction cascade, *i.e.* $J=4$ and $J=12$, respectively (due to spatial symmetries). Additionally, at each stage, we employ a single unary potential $\phi_{J+1}(\mathbf{x}_{(c)},\mathbf{y})$, where $\mathcal{C}_{J+1}$ is simply the set of all individual pixels. See Fig. 10(a) for an illustration of the neighborhood structure.

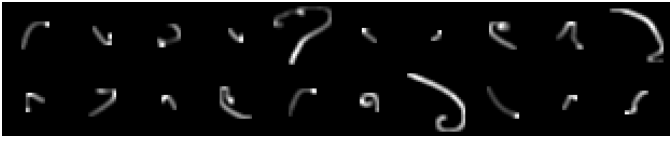

Fig. 3. **Examples of artificially generated blur kernels.**

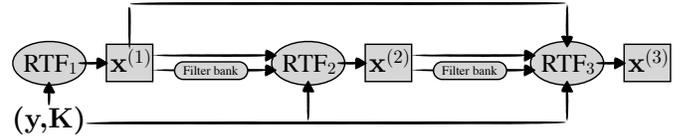

Fig. 4. **RTF prediction cascade (deblurring).** Only three stages are shown. Cascade similar for denoising, see text for details.

We extend previous RTF-based approaches to our setting by *(a)* incorporating the more general Gaussian likelihood if needed, *e.g.* for non-blind image deblurring, as outlined in Eqs. (8) and (9), and *(b)* by assembling multiple RTFs into a model cascade that iteratively refines the prediction. The cascade will be detailed in Sec. 3.3.

Note that the RTF generalizes the Gaussian CRF of Tappen *et al.* [1] in two ways: First, the potentials of an RTF are non-linearly dependent on the input image via non-parametric regression trees. Second, the model parameters of arbitrary pairwise Gaussian potentials (with full mean and covariance) are regressed from the input image, whereas [1] restrict their parameterization to diagonal weighting of filter responses.

### 3.2 Training

While probabilistic training is possible [9], we here follow [3] and learn the regressors using loss-based training, in particular, such that the average *peak signal-to-noise ratio* (PSNR) over all $N$ training images,

$$\text{psnr}(\hat{\mathbf{x}};\mathbf{x}_{\text{gt}}) = \frac{1}{N}\sum_{i=1}^{N} 20\log_{10}\left(\frac{R\sqrt{D_i}}{\|\hat{\mathbf{x}}^{(i)} - \mathbf{x}_{\text{gt}}^{(i)}\|}\right) \quad (11)$$

is maximized, where $D_i$ denotes the number of pixels in the ground truth image $\mathbf{x}_{\text{gt}}^{(i)}$ and the predicted image $\hat{\mathbf{x}}^{(i)}$ (obtained via Eq. 9), and $R$ is the maximum intensity level of a pixel (*e.g.*, $R=255$). All parameters of the model, including the split functions in the tree and the linear regressors in the leaves, are chosen to maximize PSNR [*cf.* 3].

Discriminative training necessitates a sufficient amount of training data to ensure generalization. For image denoising, it is easy to synthesize noisy versions of clean ground truth images by adding pixel-independent Gaussian noise (here using standard deviation $\sigma=25$). We use crops of $256 \times 256$ pixels from the Berkeley segmentation dataset [34] as ground truth images. Most image denoising benchmarks (including the one used in our experiments) also consist of synthesized noisy images, hence the training data matches the setting. For image deblurring, supplying appropriate training data is more challenging. Since capturing image pairs of blurred and clean images is difficult, one possible avenue is to also synthesize training data by blurring clean images with realistic blurs. Unfortunately, existing databases [13, 14] only supply a relatively limited number of blur kernels, and moreover serve also for testing. Hence the model should not be trained on these. We address this problem by generating realistic-looking blur kernels via sampling random 3D trajectories using a simple linear motion model; the obtained trajectories are projected and rasterized to random square kernel sizes in the range from $5 \times 5$ up to $27 \times 27$ pixels (see Fig. 3). While it would of course be possible to create even more realistic kernels through more accurate models of camera shake motion[4], we find that these synthetic kernels already allow to generalize well to unseen real blur (*cf.* Sec. 4). We synthetically generate blurred images by convolving each clean image with an artificially generated blur kernel, and subsequently add pixel-independent Gaussian noise (using standard deviations $\sigma=2.55$ or $0.5$, see experiments in Sec. 4). We use crops of $128 \times 128$ pixels from the training portion of the Berkeley segmentation dataset [34] as ground truth images.

### 3.3 RTF prediction cascade

**Image deblurring.** As argued in Sec. 2, it is difficult to directly regress good local image models from the blurred input image. Therefore, we employ a cascade of RTF models, where each subsequent model stage uses the output of all previous models as features for the regression (see Fig. 4 for an illustration).

We train the first stage of the cascade with minimal conditioning on the input image to avoid overfitting. Concretely, this means the parameters of the unary and pairwise potentials are only linearly regressed from the pixels in the respective cliques (plus a constant pseudo-input, *cf*. [3]); we do not use a regression tree. We further use an 8-connected graph structure, resulting in one unary and four pairwise types of potentials (horizontal, vertical, and two diagonals, *cf*. Fig. 10(a)). We train this model with 200 pairs of blurred and clean images, which is ample since there are only few model parameters. This model will be referred to as $RTF_1$.

---

[4]. We think that on average our synthetic blur kernels may in fact be more challenging than typical real ones.

While we do not expect competitive results from $RTF_1$, it is able to remove the dominant blur from the input image (*cf*. Sec. 4 and Fig. 9) and makes it much easier for subsequent RTF stages to regress good CRF potentials. Besides the blurred input image, the second stage, $RTF_2$, thus uses the output of $RTF_1$ as an input feature. We additionally evaluate a filter bank on the output of $RTF_1$ to obtain more expressive features. We therein follow [3], which obtained improved denoising results using the output of a filter bank as input to the regressor. However, we use a different filter bank here, the 16 generatively trained $5 \times 5$ filters from the recent *fields of experts* model of [35]; we found these to outperform other filter banks we have tried, including those used in [3].

We use all these features for the split tests in the regression tree (non-linear regression), as well as for the linear potential parameter regressor that is stored in each leaf of the tree. We choose regression trees of depth 7. All subsequent model stages, *i.e.* $RTF_3$, $RTF_4$, *etc.*, take as features the outputs from all previous stages, where the filter bank is always only evaluated on the directly preceding model output; see Fig. 4 for a schematic. Starting with $RTF_2$, the Gaussian CRF at each layer uses a 24-connected graph, *i.e.* each pixel is connected to all others in a $5 \times 5$ neighborhood. Due to the increased number of model parameters, we train $RTF_2$ and each subsequent stage with 500 training images, randomly cropped from the training portion of the Berkeley segmentation dataset [34] and blurred with randomly chosen artificial blur kernels (different at each stage).

**Image denoising.** Although it is much easier to directly regress good local image models from a noisy input image, image denoising can also benefit from using a model cascade, as demonstrated in our experimental evaluation (Sec. 4). However, in contrast to our deblurring cascade, we use the same RTF model architecture at each stage, in particular a 24-connected graph structure ($5\times 5$ neighborhood), filter bank responses on the output of the directly preceding model stage (or the input image for $RTF_1$), and regression trees of depth 10. We train each stage with the same 400 training images, cropped from the Berkeley segmentation dataset [34]. A minor technical difference to our deblurring cascade is that in addition to the original noisy input image, each model stage only uses the output of the directly preceding model stage (*cf*. Fig. 4) as feature for the regression (including filter bank responses thereon).

An interesting property of our model cascade in general is that it yields a restored image after every stage,

TABLE 1. Average PSNR (dB) on 68 images from [23] for image denoising with $\sigma = 25$ (not quantized); except result of [4], left part reproduced from Chen *et al*. [32]. On the right, each row shows the results from the respective stage of our model.

| Method | PSNR | Stage | PSNR |
|---|---|---|---|
| KSVD [38] | 28.28 | $RTF_1$ | 28.24 |
| $5\times 5$ FoE [35] | 28.40 | $RTF_2$ | 28.62 |
| BM3D [39] | 28.56 | $RTF_3$ | 28.70 |
| LSSC [40] | 28.70 | $RTF_4$ | 28.74 |
| EPLL [41] | 28.68 | $RTF_5$ | 28.75 |
| opt-MRF [32] | 28.66 | | |
| MLP [4] | 28.85 | | |

TABLE 2. Average PSNR (dB) on 68 images from [23] for image denoising with $\sigma = 25$ (8-bit quantized). On the right, each row shows the results from the respective stage of our model.

| Method | PSNR | Stage | PSNR |
|---|---|---|---|
| $3\times 3$ FoE [28] | 27.90 | $RTF_1$ | 28.25 |
| BLS-GSM [42] | 27.98 | $RTF_2$ | 28.61 |
| $5\times 5$ FoE [35] | 28.22 | $RTF_3$ | 28.69 |
| LSSC [40] | 28.23 | $RTF_4$ | 28.73 |
| BM3D [39] | 28.31 | $RTF_5$ | 28.74 |

not only at the end. Even if a deep cascade was trained, at test time we can trade off computational resources versus quality of the restored image by stopping after a certain stage (*cf*. Fig. 9; see [36] for a segmentation approach that can also be stopped at intermediate stages).

The cascade architecture has another advantage: because each model in the cascade has access to both the original input image as well as the output of the previous cascade stage, each stage of the cascade enlarges the learning capacity of the overall system. Our cascade architecture therefore provides *nested model classes*, as used in structural risk minimization [37].

## 4 EXPERIMENTS

### 4.1 Image denoising

We first evaluate our approach for image denoising, with a model architecture at each stage that is comparable to that of Jancsary *et al*. [3]. In contrast to [3], however, we (*a*) use a model cascade, and (*b*) choose the established denoising benchmark of 68 grayscale images from [23] (which do not contain images used for training our models). The main aim of these experiments is to demonstrate that a model cascade is beneficial, even for the (comparatively) simpler task of image denoising.

While the denoising results of [3] could not reach state-of-the-art performance without incorporating the results of other denoising methods such as BM3D [39] as features for the regression trees, our RTF prediction cascade achieves state-of-the-art performance using only the input image (and derived features via the given filter bank). Tab. 1 shows that the third model stage $RTF_3$ is already on par with the second-best competitor LSSC [40], while additional stages further improve

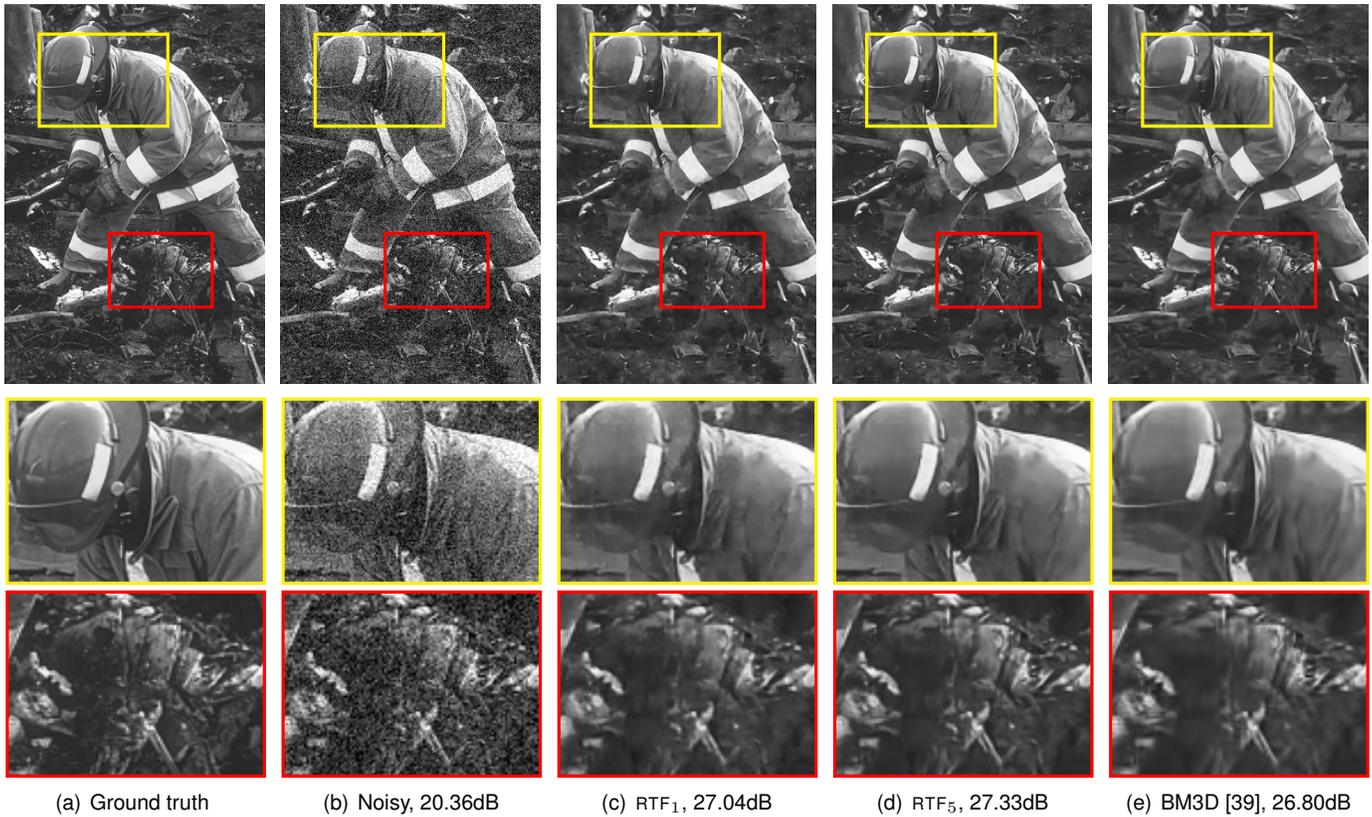

(a) Ground truth    (b) Noisy, 20.36dB    (c) $\text{RTF}_1$, 27.04dB    (d) $\text{RTF}_5$, 27.33dB    (e) BM3D [39], 26.80dB

Fig. 5. **Image denoising example (cropped).** While the result of the first stage $\text{RTF}_1$ *(c)* is already quite good, it can further be improved by additional stages of our model cascade *(d)*, both in terms of PSNR and also visually, where noise in smooth regions is further reduced (such as the firefighter's clothes), while at the same time not oversmoothing textured regions, *e.g.* the rubble at the bottom of the image (which happens for BM3D *(e)*). *Best viewed on screen.*

TABLE 3. Average PSNR (dB) on $68$ images (color versions of those used by [23]) for color image denoising with $\sigma = 25$ (added channel-independently, $8$-bit quantized).

| Model | PSNR | | | | |
|---|---|---|---|---|---|
| CBM3D [43] | 30.18 | | | | |
| Ours: | $\text{RTF}_1$ | $\text{RTF}_2$ | $\text{RTF}_3$ | $\text{RTF}_4$ | $\text{RTF}_5$ |
| Channel-independent | 28.20 | 28.55 | 28.64 | 28.67 | 28.68 |
| Channels jointly | 30.01 | **30.57** | | | |

performance marginally; the biggest performance improvement is achieved at the second stage. Our model is only outperformed by the neural network of Burger *et al.* [4], who trained a multi-layer perceptron (MLP) with millions of parameters to denoise image patches. In contrast to our model cascade, their MLP was trained on a huge database of 362 million training examples, which required about a month of training time on a GPU.

While [3] trained and tested their model without quantizing the images after adding synthetic noise, we additionally considered 8-bit quantized noisy images, *i.e.* image intensity values are rounded and range-limited, *i.e.* in $[0, \ldots, 255]$, as they would be in commonly-used image formats. Repeating the same experiment for 8-bit quantized images shows that we achieve virtually identical results (Tab. 2), while the performance of all competing methods deteriorates (often substantially, up to 0.47dB for LSSC). This highlights a strength of the RTF model, which does not make any noise assumptions[5] and can therefore easily deal with the additional quantization noise. A denoising example is shown in Fig. 5, which also compares the results of the first and last stage of our prediction cascade.

**Color image denoising.** As an additional experiment, we trained a two-stage RTF cascade for color image denoising. To that end, we use the same basic model architecture as for grayscale denoising, but with the original RGB color images from the Berkeley segmentation dataset. We do not make an attempt to use a realistic color noise model, but instead add Gaussian noise to each color channel independently (followed by 8-bit quantization). This experiment aims to show that the RTF can easily exploit correlations between the color channels, and that a model cascade is also beneficial in this case. We employ the same 68 benchmark images, but use the original color images and versions with synthetic noise, generated as described above. Compar-ing the performance of our dedicated color-denoising RTF cascade to using our grayscale-denoising RTF cascade independently for each channel (for R, G, and B color channels) reveals its superiority (Tab. 3). It outperforms the baseline grayscale model strongly by about $1.9$dB PSNR, even after only two model stages. Furthermore, we outperform the dedicated color denoising approach CBM3D [43]. Without $8$-bit quantization, CBM3D achieves a PSNR of $30.68$dB, whereas we might expect a similar performance level of our model as in the case of quantized values (*cf*. Tabs. 1 and 2). Fig. 6 shows results of our two model cascades applied to color denoising and also compares with CBM3D.

---

5. This only applies to our image denoising experiments, where we do not incorporate a likelihood component as we do for image deblurring (*cf*. Sec. 3).

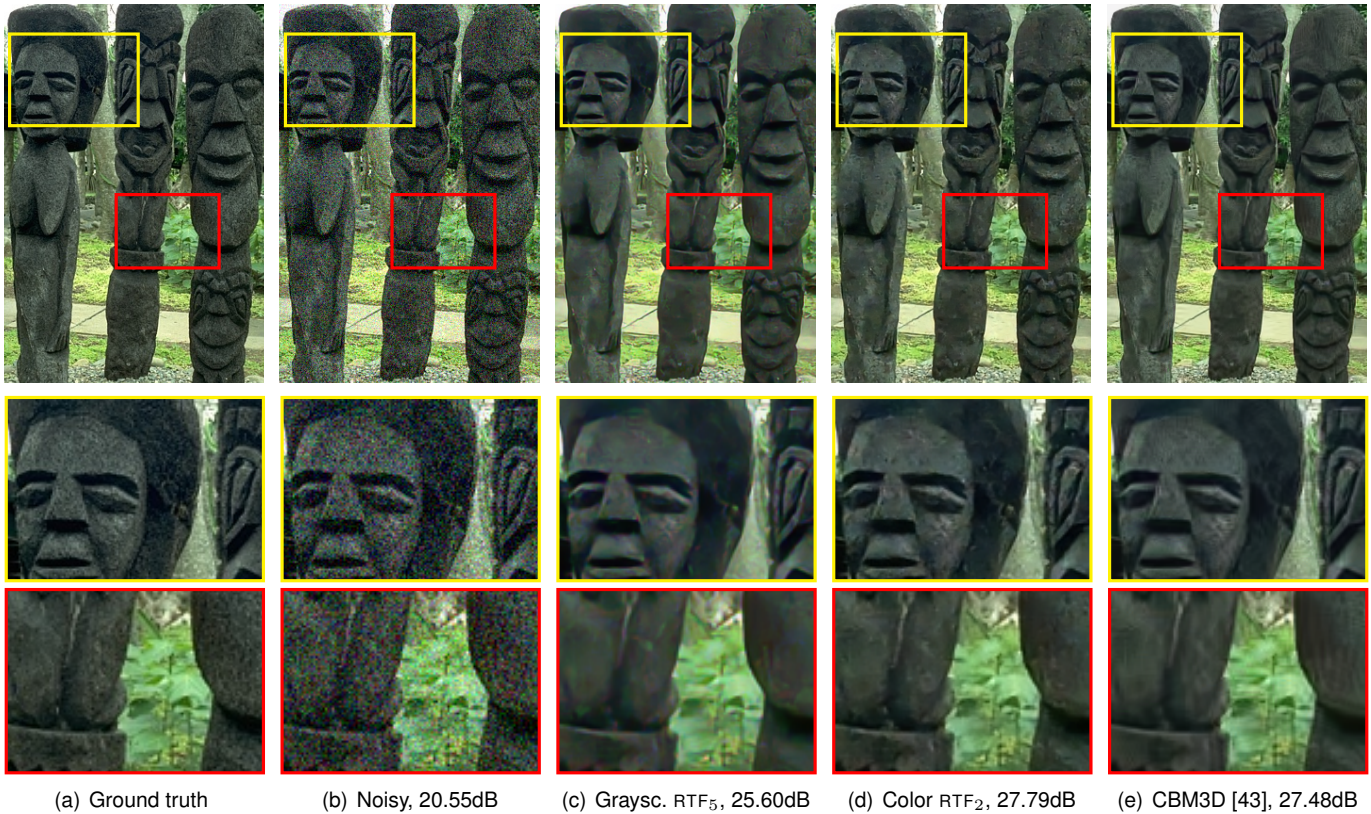

(a) Ground truth  (b) Noisy, 20.55dB  (c) Graysc. RTF$_5$, 25.60dB  (d) Color RTF$_2$, 27.79dB  (e) CBM3D [43], 27.48dB

Fig. 6. **Color denoising example (cropped).** The trained RTF cascade for color denoising *(d)* leads to better quantitative (PSNR) and qualitative results, as compared to applying a model cascade (trained for grayscale image denoising) independently for R, G, and B color channels *(c)*. Correlations between the color channels are exploited to avoid oversmoothing and color artifacts (*cf*. *(c)*). Our results *(d)* are competitive with the color denoising method CBM3D *(e)*. *Best viewed on screen.*

## 4.2 Image deblurring

To demonstrate the performance of our approach for the more difficult problem of image deblurring, we apply it to three challenging datasets, specifically to highlight individual benefits. First, we analyze the performance in the typical evaluation scenario for non-blind deblurring, *i.e.* when training and testing is carried out with (nearly) perfect blur kernels. Second, we evaluate the generalization ability of our approach by training the model to deal with imperfect blur kernels. This is important for blind deblurring, where the estimated blur kernels generally contain some errors. Third, we demonstrate the applicability to realistic higher-resolution images. Please note that images and kernels are always kept strictly separate for training and testing in all experiments.

TABLE 4. Average PSNR (dB) on $64$ images from [7] for image deblurring with two noise levels. Left half reproduced from [7] for ease of comparison.

| | $\sigma$ | | | $\sigma$ | |
|---|---|---|---|---|---|
| Method | 2.55 | 7.65 | Stage | 2.55 | 7.65 |
| Lucy-Richardson [17, 18] | 25.38 | 21.85 | RTF$_1$ | 26.33 | 24.23 |
| Krishnan and Fergus [6] | 26.97 | 24.91 | RTF$_2$ | 28.21 | 25.54 |
| Levin *et al*. [5] | 28.03 | 25.36 | RTF$_3$ | 28.50 | 25.75 |
| $5 \times 5$ FoE (MAP) [23] | 28.44 | 25.66 | RTF$_4$ | 28.58 | 25.81 |
| Pairw. MRF (MMSE) [7] | 28.24 | 25.63 | RTF$_5$ | 28.65 | 25.87 |
| $3 \times 3$ FoE (MMSE) [7] | 28.66 | 25.68 | RTF$_6$ | **28.67** | **25.89** |

**Standard evaluation.** We trained a six-stage RTF prediction cascade as described in Sec. 3 and evaluate all stages individually on $64$ test images taken from [7]. Training images have been blurred synthetically with $1\%$ additive white Gaussian noise ($\sigma = 2.55$); test images with both $\sigma = 2.55$ and a higher noise level of $\sigma = 7.65$. While we used artifical blur kernels to generate our training data, the test images from [7] have been created with the realistic kernels from [13]. The blur kernels used for deblurring in the benchmark are slightly perturbed from the ground truth to mimic kernel estimation errors (following *e.g.* [6]), but the perturbation with Gaussian noise of a small variance is somewhat minor and does not necessarily reflect typical kernel estimation errors; hence we test a more realistic scenario later on (see below). We compare our average PSNR performance to all methods that were evaluated in [7]. The results in Tab. 4 show that we achieve state-of-the-art performance that is on par with the high-quality sampling-based approach of Schmidt *et al.* [7] at $\sigma = 2.55$, and even outperforms it at $\sigma = 7.65$ despite not being trained for this noise level (only $\alpha$ was adapted, see Eq. 8). As we shall discuss below, our approach is many times faster, however. At the noise level our model is trained for ($\sigma = 2.55$), we strongly outperform the efficient half-quadratic regularization approach of Krishnan and Fergus [6] by over 1.5dB, and the popular method of Levin *et al.* [5] by 0.6dB. The clear performance gains at the higher noise level demonstrate our model's noise generalization. We further notice that the weakly conditional first stage (RTF$_1$) leads only to modest performance levels here; RTF$_2$ and RTF$_3$ boost the performance substantially further. Later stages lead to additional gains, but less so. Aside from the raw numbers, it is noteworthy that our model is able to preserve small details, while at the same time reconstructing smooth areas well (see Fig. 7 for an example). Note that this is not the case for the approaches tested in [7].

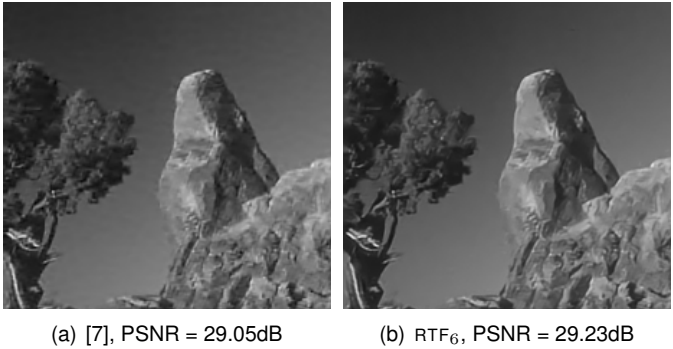

(a) [7], PSNR = 29.05dB  (b) RTF$_6$, PSNR = 29.23dB

Fig. 7. **Deblurring example (cropped).** Qualitative comparison with the high-quality approach of [7] ($3 \times 3$ FoE, MMSE estimation). Our approach *(b)* reconstructs smooth and textured areas well, exhibits fewer artifacts, and is many times faster. *Best viewed zoomed in on screen.*

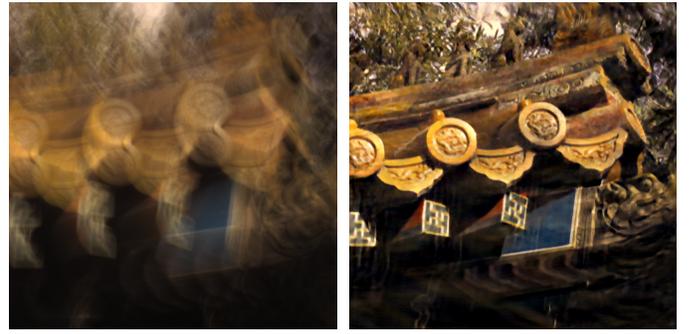

Fig. 8. Deblurring example from the benchmark of [14] (*cf*. Tab. 6), showing the result of our RTF$_2$ model *(right)* given the blurred image *(left)* and the kernel estimates by [21].

This demonstrates that when testing (and training) is done with the correct (*i.e.* ground truth) blur kernels, our approach achieves very good results. Even though we train our model on artificially generated blur kernels (Fig. 3), it apparently generalizes well to real blurs.

**Adaptation to kernel estimation errors.** Blind deblurring approaches often produce kernel estimates with substantial errors, which can cause ringing artifacts in the restored image [*cf*. 44]. Hence, it is important to evaluate and adapt our model to this realistic scenario. To train our model for this task, we experimented with adding noise to the ground truth kernels and also used estimated kernels for training.

We consider the data of Levin *et al.* [15] as a benchmark, which provides several kernel estimates besides blurred and ground truth images for 32 test instances, as well as deblurring results with the various kernel estimates. Since the amount of noise in these blurred images is significantly lower than in the benchmark of [7], we only added Gaussian noise with $\sigma = 0.5$ to our training images. We evaluate average PSNR performance over all 32 images (using code by [15] to account for translations in kernel estimates) instead of error ratios as in [15], since we are not interested in the quality of the estimated kernels itself, but rather the final restoration performance given the estimated kernels.

The results in Tab. 5 show that training with ground truth kernels leads to subpar performance when kernel estimates are used at test time. Adding noise to the ground truth kernels for training leads to improved results of RTF$_1$ with estimated kernels at test time, but performance of our second stage model RTF$_2$ already deteriorates; hence those noisy kernels are not an ideal proxy for real kernel estimates. However, we achieve superior results by training our model with a mix of perfect and estimated kernels (obtained with the method of Xu and Jia [21]), *i.e.* for half of the synthetically blurred training images we use an estimated kernel instead of the ground truth kernel[6]. Compared to the deblurred images from [15] (which used the non-blind approach of [5]), we achieve substantial performance improvements for deblurring with estimated kernels of up to 0.72dB (for kernels from [19]). Furthermore, it is interesting to note that the first stage of our model already achieves good performance; this is presumably due to the much reduced amount of noise in this benchmark[7].

**Runtime.** The computational demand of our method is comparable to the half-quadratic approach of [5], but uses this computational budget more effectively due to its discriminative nature (*cf*. Sec. 2 and Fig. 2). Also note that the tree-based regressor is very efficient. As a result, we achieve state-of-the-art performance on par with the best result of [7], but much faster: about $2$ seconds per image in Tab. 4 (all six model stages combined) compared to $4$ minutes for [7]. For the benchmark in Tab. 5 with larger images, we require around $3$ seconds for each model stage.

---

6. Here, we trained RTF$_1$ and RTF$_2$ with the same 200 images as it was time-consuming to obtain good enough kernel estimates for training.

7. Theoretically, in the absence of noise, non-blind deblurring can be solved exactly without any regularization by inverting the blur matrix.

TABLE 5. Deblurring results (PSNR in dB, average over $32$ images from [15]) that analyze the ability to cope with kernel estimation errors. The kernel estimates of [15, 19, 20] are provided by [15]; the kernel estimates of [21] are obtained using the authors' code. The last row shows the average performance of the non-blind method of [5] for various kernels, as provided by [15]. For the kernel estimates of [15] ($4^{th}$ column), we used the "free energy with diagonal covariance approximation" algorithm in the filter domain.

| Method | Kernels for training | Kernels for testing | | | | | |
|---|---|---|---|---|---|---|---|
| | | GT | Levin et al. [15] | Cho and Lee [20] | Fergus et al. [19] | GT + Noise | Xu and Jia [21] |
| $\text{RTF}_1$ | GT | 32.76 | 29.41 | 28.29 | 27.86 | 26.67 | 29.04 |
| $\text{RTF}_2$ | GT | 33.81 | 29.52 | 27.76 | 27.84 | 26.52 | 28.29 |
| $\text{RTF}_1$ | GT + Noise | 32.08 | 29.73 | 29.36 | 28.49 | 28.69 | 30.25 |
| $\text{RTF}_2$ | GT + Noise | 30.51 | 29.03 | 28.75 | 27.58 | 30.34 | 29.56 |
| $\text{RTF}_1$ | Mix of GT & [21] | 32.90 | 29.90 | 29.33 | 28.63 | 28.10 | 30.30 |
| $\text{RTF}_2$ | Mix of GT & [21] | 33.97 | 30.40 | 29.73 | 29.10 | 28.07 | 30.84 |
| [5] | — | 32.73 | 30.05 | $29.71^8$ | 28.38 | — | — |

TABLE 6. Performance gain (PSNR in dB) over the results of Xu and Jia [21] in the benchmark of Köhler et al. [14] for each combination of $4$ test images and $12$ blur kernels. We use the provided blur kernel estimates of [21] with our $\text{RTF}_2$ model for non-blind deblurring. We can improve the performance in $43$ of $48$ test instances, on average about $0.41$dB.

| | Image 1 | Image 2 | Image 3 | Image 4 |
|---|---|---|---|---|
| Kernel 1 | +0.44 | +0.54 | +1.05 | +0.76 |
| Kernel 2 | +0.44 | +0.27 | +0.38 | +0.46 |
| Kernel 3 | +0.02 | +0.03 | +0.39 | −0.26 |
| Kernel 4 | +0.31 | +0.30 | +0.61 | +0.27 |
| Kernel 5 | +0.61 | +0.44 | +0.64 | +0.05 |
| Kernel 6 | +0.40 | +0.41 | +1.03 | +0.48 |
| Kernel 7 | +0.24 | +0.55 | +0.45 | +0.31 |
| Kernel 8 | +0.76 | +0.56 | +2.17 | +1.73 |
| Kernel 9 | +0.35 | −0.09 | +0.02 | +0.23 |
| Kernel 10 | +0.19 | −0.55 | +0.25 | +0.29 |
| Kernel 11 | −0.19 | −0.43 | +0.46 | +0.09 |
| Kernel 12 | +0.76 | +0.04 | +0.66 | +0.64 |

**Realistic higher-resolution images.** We consider the recent benchmark for camera shake by Köhler et al. [14] to demonstrate results on realistic higher-resolution images; these images may substantially violate our model's stationary blur and Gaussian noise assumptions (which can deteriorate performance [*cf*. 45, 46]). The benchmark consists of $4$ color images of size $800 \times 800$ pixels blurred by $12$ different real camera motions, yielding $48$ images in total. The overall best performing blind deblurring approach in this benchmark is the one by Xu and Jia

8. This result taken from [15] may have employed the non-blind method from [20].

[21] despite making a stationary blur assumption, *i.e.* the same blur kernel is used in all parts of the image. We use the provided kernel estimates by [21] from the benchmark dataset, but with our non-blind method to obtain the deblurred image (treating color channels R, G, and B independently). Tab. 6 shows that performance (evaluated using the provided code) can substantially be improved by using our $\text{RTF}_2$ model instead of their non-blind step (which is related to [6]). While Xu and Jia's non-blind step is inherently faster, it does lead to substantially worse results, here on average $0.41$dB. Fig. 8 shows an example of a deblurred image. Note that the $\text{RTF}_2$ model used here is the same as in Tab. 5, *i.e.* trained with a mix of ground truth and estimated kernels (using [21]), and additive Gaussian noise with $\sigma = 0.5$.

## 5 DISCUSSION

### 5.1 Training dataset

For image denoising (Tabs. 1 and 2) and image deblurring in typical evaluation scenarios (*i.e.* true blur kernel and noise level known at test time, Tab. 4), we have trained RTF model cascades for up to six stages, with each additional stage improving restoration performance (although with diminishing improvements in the later stages). However, this does not apply to deblurring in the context of blind deblurring, *i.e.* where erroneous estimated blur kernels are used at test time. Especially under realistic conditions (Tab. 6), the blur might be spatially varying and the noise may not be Gaussian. Under these conditions, it is much more difficult to find a suitable training set in a discriminative setting such as ours. We initially tried using noisy blur kernels as a proxy for estimated kernels at test time, but only achieved performance improvements at the first model stage (*cf*. Tab. 5); in fact it was challenging to learn a second model stage that would improve upon the first. While we showed it to be possible to outperform existing approaches by training our model with a mix of ground truth and estimated kernels (*cf*. Sec. 4.2), we believe substantially improved results could be achieved with training datasets that more closely match the conditions encountered at test time. Future work should thus aim to provide realistic data with ground truth also for training discriminative approaches, not only for benchmarking.

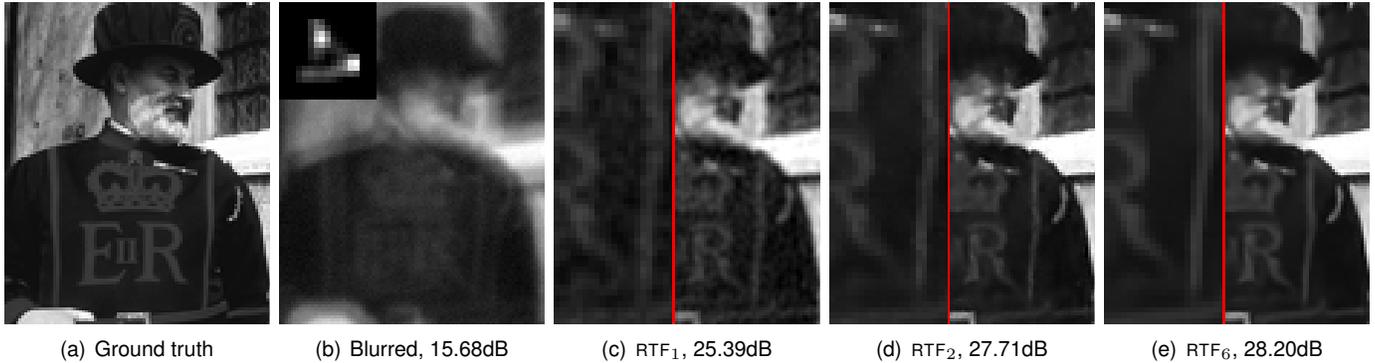

(a) Ground truth  (b) Blurred, 15.68dB  (c) RTF$_1$, 25.39dB  (d) RTF$_2$, 27.71dB  (e) RTF$_6$, 28.20dB

Fig. 9. **Deblurring example at different model stages.** The first stage RTF$_1$ removes dominant blur from the image *(c)*, but artifacts remain. The second stage RTF$_2$ *(d)* substantially improves upon this result quantitatively (PSNR in dB) and qualitatively. Further model stages continue to suppress noise and refine image details *(e)*. The left sides of *(c–e)* show a closeup view of image details on the respective right sides. The blur kernel is shown at the upper left of *(b)*, scaled and resized for better visualization. *Best viewed on screen.*

## 5.2 Model connectivity and comparison

Random field models for image restoration typically use (manually defined) pairwise connectivity (4-connected neighborhood, *i.e.* horizontal and vertical direct neighbor), or alternatively follow the fields of experts (FoE) framework [23], which models responses of a (learned) filter bank of extended size (5×5 often used, see Fig. 10(b) for an example). In contrast, the regression tree field, as introduced by Jancsary *et al.* [9] and also used here, employs learned (and possibly long-range) pairwise connections; see Fig. 10(a) for an illustration. In a Gaussian random field, such as the RTF, all high-order factors can always be expressed through pairwise ones [30]. Hence, no modeling power is lost by restricting factors to pairwise (and unary) connectivity.

In Fig. 10, both RTF and fields of experts are shown with 8-connected neighborhoods, *i.e.* the central pixel is connected to its nearest 8 neighbors (depicted in dark gray). We have used a 24-connected neighborhood in most RTF model stages. An identical connectivity is achieved via a fields of experts model with 3×3 filters. In general, an FoE model with filters of size $m \times m$ yields a $2m^2 - 1$ neighborhood connectivity. In an RTF, large connectivity can be achieved by adding more long-range pairwise connections, but this becomes prohibitively expensive to train in the current setting, where training complexity is linear in the number of factor types. Of course, one could modify the RTF to also model filter responses, which may be the subject of future work.

## 6  SUMMARY AND CONCLUSIONS

From a novel analysis of common half-quadratic regularization, we introduced a discriminative image restoration approach, applicable to image restoration problems that can be expressed through (arbitrary) quadratic data terms. We enable discriminative prediction in the context of challenging Gaussian image corruption models by separating the instance-specific parameters of the data model from the discriminative parameter regression, which for deblurring allows coping with arbitrary blur kernels at test time without needing to retrain the model. Moreover, a discriminative prediction cascade helps to overcome the problem of regressing suitable parameters directly from the input image. Our proposed cascade model is based on regression tree fields at each stage, which are trained by loss minimization on training data generated according to the given data term.

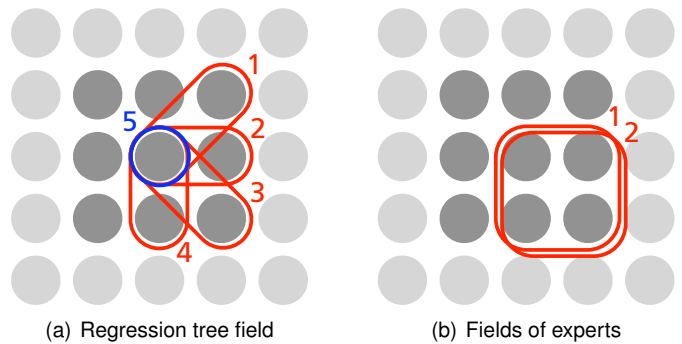

(a) Regression tree field    (b) Fields of experts

Fig. 10. **Factor types for 8-connected random fields (shown anchored at central pixel).** *(a)* RTF with four pairwise (red) and one unary (blue) factor type, and *(b)* filter-based random field model (fields of experts [23]) with two filters of size 2×2 (red).

We demonstrated its merit for image denoising and especially for the problem of non-blind deblurring. For deblurring, we employed synthesized blur kernels to generate training data. We demonstrated state-of-the art performance on several challenging benchmarks, including robustness to kernel estimation errors in the context of blind deblurring. Our approach is not limited to image denoising and deblurring, and can be extended to other image restoration applications, especially when their data term takes a quadratic form.

## ACKNOWLEDGMENTS

We thank Pushmeet Kohli for suggesting the topic of discriminative deblurring using a non-parametric model like the RTF; we also thank Stephan Richter for help in preparation of Fig. 2. Parts of this work have been done when Uwe Schmidt interned at Microsoft Research Cambridge, UK. Funding is partly provided by Microsoft Research through its PhD Scholarship Programme, as well as from the European Research Council under the European Union's Seventh Framework Programme (FP/2007-2013) / ERC Grant Agreement n. 307942.

## REFERENCES

[1] M. Tappen, C. Liu, E. H. Adelson, and W. T. Freeman, "Learning Gaussian conditional random fields for low-level vision," in *IEEE Conf. on Comp. Vis. and Pat. Recog. (CVPR)*, Jun. 2007.

[2] A. Barbu, "Learning real-time MRF inference for image denoising," in *IEEE Conf. on Comp. Vis. and Pat. Recog. (CVPR)*, Jun. 2009.

[3] J. Jancsary, S. Nowozin, and C. Rother, "Loss-specific training of non-parametric image restoration models: A new state of the art," in *Eur. Conf. on Comp. Vis. (ECCV)*, ser. Lect. Notes in Comp. Sci., A. Fitzgibbon, S. Lazebnik, P. Perona, Y. Sato, and C. Schmid, Eds. Springer, 2012, vol. 7578.

[4] H. C. Burger, C. J. Schuler, and S. Harmeling, "Image denoising: Can plain neural networks compete with BM3D?" in *IEEE Conf. on Comp. Vis. and Pat. Recog. (CVPR)*, Jun. 2012.

[5] A. Levin, R. Fergus, F. Durand, and W. T. Freeman, "Image and depth from a conventional camera with a coded aperture," *ACM Transactions on Graphics*, vol. 26, no. 3, pp. 70:1–70:9, Jul. 2007.

[6] D. Krishnan and R. Fergus, "Fast image deconvolution using hyper-Laplacian priors," in *Adv. in Neur. Inf. Proc. Sys. (NIPS)*, vol. 22, 2009.

[7] U. Schmidt, K. Schelten, and S. Roth, "Bayesian deblurring with integrated noise estimation," in *IEEE Conf. on Comp. Vis. and Pat. Recog. (CVPR)*, Jun. 2011.

[8] C. J. Schuler, H. C. Burger, S. Harmeling, and B. Schölkopf, "A machine learning approach for non-blind image deconvolution," in *IEEE Conf. on Comp. Vis. and Pat. Recog. (CVPR)*, Jun. 2013.

[9] J. Jancsary, S. Nowozin, T. Sharp, and C. Rother, "Regression tree fields – an efficient, non-parametric approach to image labeling problems," in *IEEE Conf. on Comp. Vis. and Pat. Recog. (CVPR)*, Jun. 2012.

[10] D. Geman and C. Yang, "Nonlinear image recovery with half-quadratic regularization," *IEEE Trans. Image Process.*, vol. 4, no. 7, pp. 932–946, Jul. 1995.

[11] D. Geman and G. Reynolds, "Constrained restoration and the recovery of discontinuities," *IEEE Trans. Pattern Anal. Mach. Intell.*, vol. 14, no. 3, pp. 367–383, Mar. 1992.

[12] P. Charbonnier, L. Blanc-Féraud, G. Aubert, and M. Barlaud, "Two deterministic half-quadratic regularization algorithms for computed imaging," in *IEEE Int. Conf. on Image Proc. (ICIP)*, vol. 2, Nov. 1994.

[13] A. Levin, Y. Weiss, F. Durand, and W. T. Freeman, "Understanding and evaluating blind deconvolution algorithms," in *IEEE Conf. on Comp. Vis. and Pat. Recog. (CVPR)*, Jun. 2009.

[14] R. Köhler, M. Hirsch, B. Mohler, B. Schölkopf, and S. Harmeling, "Recording and playback of camera shake: Benchmarking blind deconvolution with a real-world database," in *Eur. Conf. on Comp. Vis. (ECCV)*, ser. Lect. Notes in Comp. Sci., A. Fitzgibbon, S. Lazebnik, P. Perona, Y. Sato, and C. Schmid, Eds., vol. 7578. Springer, 2012.

[15] A. Levin, Y. Weiss, F. Durand, and W. T. Freeman, "Efficient marginal likelihood optimization in blind deconvolution," in *IEEE Conf. on Comp. Vis. and Pat. Recog. (CVPR)*, Jun. 2011.

[16] U. Schmidt, C. Rother, S. Nowozin, J. Jancsary, and S. Roth, "Discriminative non-blind deblurring," in *IEEE Conf. on Comp. Vis. and Pat. Recog. (CVPR)*, Jun. 2013.

[17] L. B. Lucy, "An iterative technique for the rectification of observed distributions," *The Astronomical Journal*, vol. 79, no. 6, 1974.

[18] W. Richardson, "Bayesian-based iterative method of image restoration," *J. Opt. Soc. America*, vol. 62, no. 1, pp. 55–59, 1972.

[19] R. Fergus, B. Singh, A. Hertzmann, S. T. Roweis, and W. T. Freeman, "Removing camera shake from a single photograph," *ACM Trans. Graphics (Proc. ACM SIGGRAPH)*, vol. 25, no. 3, pp. 787–794, Jul.-Aug. 2006.

[20] S. Cho and S. Lee, "Fast motion deblurring," *ACM Trans. Graphics*, vol. 28, no. 5, 2009.

[21] L. Xu and J. Jia, "Two-phase kernel estimation for robust motion deblurring," in *Eur. Conf. on Comp. Vis. (ECCV)*, ser. Lect. Notes in Comp. Sci., K. Daniilidis, P. Maragos, and N. Paragios, Eds. Springer, 2010, vol. 6311.

[22] O. Whyte, J. Sivic, A. Zisserman, and J. Ponce, "Non-uniform deblurring for shaken images," in *IEEE Conf. on Comp. Vis. and Pat. Recog. (CVPR)*, Jun. 2010.

[23] S. Roth and M. J. Black, "Fields of experts," *Int. J. Comput. Vision*, vol. 82, no. 2, pp. 205–229, Apr. 2009.

[24] J. A. Palmer, D. P. Wipf, K. Kreutz-Delgado, and B. D. Rao, "Variational EM algorithms for non-Gaussian latent variable models," in *Adv. in Neur. Inf. Proc. Sys. (NIPS)*, vol. 18, 2006.

[25] F. Champagnat and J. Idier, "A connection between half-quadratic criteria and EM algorithms," *IEEE Signal Processing Letters*, vol. 11, no. 9, pp. 709–712, Sep. 2004.

[26] M. J. Black and A. Rangarajan, "On the unification of line processes, outlier rejection, and robust statistics with applications in early vision," *Int. J. Comput. Vision*, vol. 19, no. 1, pp. 57–91, Jul. 1996.

[27] M. J. Wainwright and E. P. Simoncelli, "Scale mixtures of Gaussians and the statistics of natural images," in *Adv. in Neur. Inf. Proc. Sys. (NIPS)*, vol. 12, 2000.

[28] U. Schmidt, Q. Gao, and S. Roth, "A generative perspective on MRFs in low-level vision," in *IEEE Conf. on Comp. Vis. and Pat. Recog. (CVPR)*, Jun. 2010.

[29] S. D. Babacan, R. Molina, M. N. Do, and A. K. Katsaggelos, "Bayesian blind deconvolution with general sparse image priors," in *Eur. Conf. on Comp. Vis. (ECCV)*, ser. Lect. Notes in Comp. Sci., A. Fitzgibbon, S. Lazebnik, P. Perona, Y. Sato, and C. Schmid, Eds., vol. 7577. Springer, 2012.

[30] M. J. Wainwright and M. I. Jordan, "Graphical models, exponential families, and variational inference," *Foundations and Trends in Machine Learning*, vol. 1, no. 1–2, 2008.

[31] Z. Tu, "Auto-context and its application to high-level vision tasks," in *IEEE Conf. on Comp. Vis. and Pat. Recog. (CVPR)*, Jun. 2008.

[32] Y. Chen, T. Pock, R. Ranftl, and H. Bischof, "Revisiting loss-specific training of filter-based MRFs for image restoration," in *Germ. Conf. Pat. Recog. (GCPR)*, ser. Lect. Notes in Comp. Sci. Springer, 2013.

[33] T. S. Cho, N. Joshi, C. L. Zitnick, S. B. Kang, R. Szeliski, and W. T. Freeman, "A content-aware image prior," in *IEEE Conf. on Comp. Vis. and Pat. Recog. (CVPR)*, Jun. 2010.

[34] P. Arbelaez, M. Maire, C. Fowlkes, and J. Malik, "Contour detection and hierarchical image segmentation," *IEEE Trans. Pattern Anal. Mach. Intell.*, vol. 33, no. 5, pp. 898–916, 2011.

[35] Q. Gao and S. Roth, "How well do filter-based MRFs model natural images?" in *Pat. Recog., Proc. DAGM-Symp.*,

ser. Lect. Notes in Comp. Sci. Springer, 2012.
[36] B. Fröhlich, E. Rodner, and J. Denzler, "As time goes by—anytime semantic segmentation with iterative context forests," in *Pat. Recog., Proc. DAGM-Symp.*, ser. Lect. Notes in Comp. Sci. Springer, 2012.
[37] V. Vapnik and A. Chervonenkis, *Theory of pattern recognition (in Russian)*. Nauka, Moscow, 1974.
[38] M. Elad, B. Matalon, and M. Zibulevsky, "Image denoising with shrinkage and redundant representations," in *IEEE Conf. on Comp. Vis. and Pat. Recog. (CVPR)*, Jun. 2006.
[39] K. Dabov, A. Foi, V. Katkovnik, and K. Egiazarian, "Image denoising by sparse 3-D transform-domain collaborative filtering," *IEEE Trans. Image Process.*, vol. 16, no. 8, pp. 2080–2095, Aug. 2007.
[40] J. Mairal, F. Bach, J. Ponce, G. Sapiro, and A. Zisserman, "Non-local sparse models for image restoration," in *IEEE Int. Conf. on Comp. Vis. (ICCV)*, Oct. 2009.
[41] D. Zoran and Y. Weiss, "From learning models of natural image patches to whole image restoration," in *IEEE Int. Conf. on Comp. Vis. (ICCV)*, 2011.
[42] J. Portilla, V. Strela, M. J. Wainwright, and E. P. Simoncelli, "Image denoising using scale mixtures of Gaussians in the wavelet domain," *IEEE Trans. Image Process.*, vol. 12, no. 11, pp. 1338–1351, Nov. 2003.
[43] K. Dabov, A. Foi, V. Katkovnik, and K. Egiazarian, "Color image denoising via sparse 3D collaborative filtering with grouping constraint in luminance-chrominance space," in *IEEE Int. Conf. on Image Proc. (ICIP)*, Oct. 2007.
[44] L. Yuan, J. Sun, L. Quan, and H.-Y. Shum, "Progressive inter-scale and intra-scale non-blind image deconvolution," *ACM Trans. Graphics*, vol. 27, no. 3, 2008.
[45] S. Cho, J. Wang, and S. Lee, "Handling outliers in non-blind image deconvolution," in *IEEE Int. Conf. on Comp. Vis. (ICCV)*, Nov. 2011.
[46] Y.-W. Tai and S. Lin, "Motion-aware noise filtering for deblurring of noisy and blurry images," in *IEEE Conf. on Comp. Vis. and Pat. Recog. (CVPR)*, Jun. 2012.